\documentclass{article} 
\usepackage{iclr2027_conference,times}
\usepackage{amsmath,amssymb}
\usepackage{mathrsfs}
\usepackage{booktabs}
\usepackage{graphicx}
\usepackage{flafter}
\usepackage{wrapfig}

\usepackage{amsmath,amsfonts,bm}

\def\eqref#1{equation~\ref{#1}}

\def\1{\bm{1}}

\DeclareMathAlphabet{\mathsfit}{\encodingdefault}{\sfdefault}{m}{sl}
\SetMathAlphabet{\mathsfit}{bold}{\encodingdefault}{\sfdefault}{bx}{n}

\usepackage{hyperref}
\usepackage{url}

\newcommand{\MVMImageNetThirtyTwoFID}{1.93}
\newcommand{\MVMImageNetThirtyTwoNFE}{65}
\newcommand{\MVMImageNetTwoFiftySixFID}{2.07}
\newcommand{\MVMImageNetTwoFiftySixNFE}{90}

\title{Mean Velocity Matching: Rethinking Generative Dynamics in Diffusion Models}

\author{%
	\textbf{Yunhong Zhang}\textsuperscript{1} \quad
	\textbf{Changjie Cao}\textsuperscript{1,}\thanks{Corresponding author.} \quad
	\textbf{Zhihua Zhang}\textsuperscript{2} \quad
	\textbf{Bingli Liu}\textsuperscript{1} \\
	\textbf{Zongjie Cao}\textsuperscript{3} \quad
	\textbf{Zongyong Cui}\textsuperscript{3} \quad
	\textbf{Ying Yang}\textsuperscript{1} \\
	\\
	\textsuperscript{1}Chengdu University of Technology, Chengdu, China \\
	\textsuperscript{2}Peking University, Beijing, China\\
	\textsuperscript{3}University of Electronic Science and Technology of China, Chengdu, China	
}
\iclrfinalcopy 
\begin{document}

	\maketitle
\fancyhead{}
\renewcommand{\headrulewidth}{0pt}	
	\begin{abstract}
		This work studies prediction parameterization for stochastic generative dynamics in diffusion models. Existing velocity-based generative models provide the simplicity of learning a single transport field, but their standard formulation is deterministic, whereas stochastic extensions generally require additional score information or an intermediate velocity-to-score reconstruction. To retain single-field prediction while directly supporting stochastic reverse dynamics, this paper introduces Mean Velocity Matching (MVM). MVM constructs a Gaussian perturbation process for which the conditional expectation of a restoration-oriented velocity, $(x_0-x_t)/t$, directly forms the reverse-SDE drift. Consequently, a single learned field is sufficient to parameterize the stochastic reverse process without separately estimating or reconstructing the score. Because direct regression of this velocity becomes unbounded near $t=0$, MVM further introduces a $\sqrt{t}$-scaled parameterization that preserves the reverse dynamics while yielding a bounded training target. The same learned field also induces a deterministic probability-flow ODE, enabling stochastic and deterministic sampling to be studied within a unified formulation. Experiments with Transformer-based generative models achieve an FID of $\MVMImageNetThirtyTwoFID$ at \MVMImageNetThirtyTwoNFE\ NFE on ImageNet $32\times32$ and $\MVMImageNetTwoFiftySixFID$ at \MVMImageNetTwoFiftySixNFE\ NFE on ImageNet $256\times256$. Controlled SDE--ODE comparisons further show that the ODE performs better under very low NFE, whereas the stochastic reverse process achieves lower FID when sufficient function evaluations are available. These results demonstrate that MVM provides a direct single-field parameterization of stochastic reverse dynamics while maintaining competitive generation quality.
	\end{abstract}

	\section{INTRODUCTION}
	
	Diffusion models have achieved strong performance in image synthesis
	\citep{1,2}, video generation \citep{3}, three-dimensional content
	generation \citep{4}, and large-scale text-to-image synthesis
	\citep{17,18,22}. They combine a gradual perturbation process with a
	learned reverse process that transports tractable noise back to the
	data distribution, and this view extends naturally from discrete-time
	chains to continuous-time stochastic differential equations (SDEs)
	\citep{5}. Transformer backbones such as DiT \citep{44} further show
	that diffusion models scale effectively with model capacity and
	computation.
	
	\begin{figure}[!t]
		\centering
		\refstepcounter{figure}\label{fig:mvm-overview}
		\begin{minipage}[t]{0.705\linewidth}
			\vspace{0pt}
			\centering
			\includegraphics[
			width=\linewidth,
			pagebox=cropbox
			]{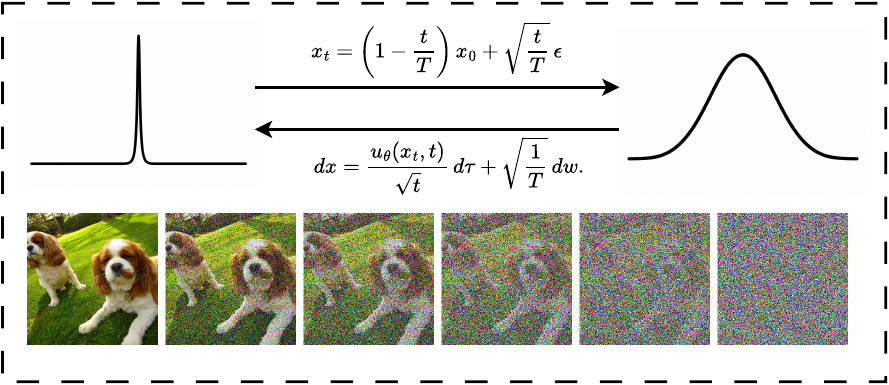}
		\end{minipage}\hfill
		\begin{minipage}[t]{0.265\linewidth}
			\vspace{0pt}
			\small
			\raggedright
			\textbf{Figure~\thefigure: Overview of Mean Velocity Matching.}
			The forward process continuously transforms data into Gaussian noise,
			whereas the learned reverse-time SDE transports noise back to the data
			distribution through mean-velocity prediction.\par
		\end{minipage}
		\vspace{-1em}
	\end{figure}
	
	A central design choice in diffusion models is the variable predicted by
	the neural network. DDPM \citep{2} adopts perturbation-noise prediction
	and reported that direct $x_0$-prediction produced worse sample quality
	in its early experiments. Later work showed, however, that the preferred
	parameterization can depend strongly on the sampling regime. In
	progressive distillation, for example, the implied $x_0$ estimate from
	$\epsilon$-prediction becomes ill-conditioned as the signal-to-noise
	ratio approaches zero, motivating direct data prediction and diffusion
	$v$-prediction \citep{25}. The latter combines clean-data and noise
	information in a single target,
	\begin{equation}
		v_t=\alpha_t\epsilon-\sigma_t x_0.
	\end{equation}
	These observations suggest that a useful prediction target may benefit
	from representing data- and noise-related information jointly rather
	than committing exclusively to either endpoint.
	
	Flow-based generative models provide a related perspective. Flow
	Matching \citep{8} and Rectified Flow \citep{10} learn a single
	velocity field that transports probability between endpoint
	distributions. Under commonly used linear interpolations between data
	and Gaussian noise, the conditional velocity is determined jointly by
	the two endpoints. Once learned, the marginal velocity directly defines
	a deterministic ordinary differential equation,
	\begin{equation}
		dx=v_\theta(x,t)\,dt,
	\end{equation}
	so generation requires only one learned field.
	
	Deterministic transport, however, is not the only possible reverse
	dynamics. Stochastic Interpolants \citep{9} show that velocity fields,
	scores, deterministic flows, and stochastic diffusions can be related
	within a common framework. Stochastic samplers can also be constructed
	from pretrained deterministic flow models. In a general construction,
	the SDE drift combines the learned flow field with score information;
	for Gaussian flow models, the score can be recovered analytically from
	the learned velocity, avoiding a separately trained score network
	\citep{51}. Nevertheless, the stochastic drift in such constructions is
	obtained by augmenting the transport velocity with, or reconstructing
	it through, score information.
	
	This paper investigates a different possibility: \emph{can a stochastic
		reverse process retain the simplicity of single-field velocity
		prediction while avoiding an intermediate velocity-to-score
		reconstruction?} To address this question, the present study introduces
	\emph{Mean Velocity Matching} (MVM), a continuous-time generative
	framework in which one learned field directly parameterizes the
	reverse-SDE drift.
	
	MVM constructs the Gaussian perturbation path
	\begin{equation}
		x_t
		=
		\left(1-\frac{t}{T}\right)x_0
		+
		\sqrt{\frac{t}{T}}\,\epsilon,
		\qquad
		\epsilon\sim\mathcal{N}(0,I),
	\end{equation}
	which continuously maps the data distribution to standard Gaussian
	noise at the terminal time $T$. For a paired forward sample
	$(x_0,x_t)$, define
	\begin{equation}
		v(x_t,t;x_0)
		=
		\frac{x_0-x_t}{t}.
	\end{equation}
	By deriving the reverse probability evolution of this process, the
	present study shows that the marginal reverse drift is
	\begin{equation}
		v^{*}(x_t,t)
		=
		\mathbb{E}
		\left[
		\left.
		\frac{x_0-x_t}{t}
		\right|
		x_t
		\right].
	\end{equation}
	Consequently, the reverse stochastic dynamics take the simple form
	\begin{equation}
		dx
		=
		v^{*}(x_t,t)\,d\tau
		+
		\sqrt{\frac{1}{T}}\,dw.
	\end{equation}
	The key property is that a single predicted field determines the
	reverse-SDE drift directly, without requiring a separately learned score
	or an intermediate velocity-to-score conversion during generation. The conditional drift is tractable to learn because squared-error
	regression on the sample-wise target $(x_0-x_t)/t$ has the same
	population minimizer as regression to its conditional expectation.
	Direct regression becomes ill-conditioned near $t=0$, where the target
	grows as $\mathcal{O}(t^{-1/2})$. MVM therefore uses the scaled
	parameterization
	\begin{equation}
		u(x_t,t)
		=
		\sqrt{t}\,v(x_t,t)
		=
		\frac{x_0-x_t}{\sqrt{t}}
		=
		\frac{\sqrt{t}}{T}x_0
		-
		\frac{\epsilon}{\sqrt{T}},
	\end{equation}
	which remains bounded near the data endpoint. During sampling, the
	reverse drift is recovered by
	$v_\theta(x_t,t)=u_\theta(x_t,t)/\sqrt{t}$. The same learned field also
	admits a deterministic probability-flow ODE with the same marginal
	distributions, enabling stochastic and deterministic sampling within a
	common formulation.
	
	The main contributions of this paper are as follows.
	\begin{itemize}
		\item
		The present study introduces MVM, in which a single learned
		velocity-like field directly parameterizes the reverse-SDE drift,
		without requiring a separately learned score field or an
		intermediate velocity-to-score reconstruction during generation.
		
		\item
		The paper derives a Gaussian perturbation process for which the
		marginal reverse drift is exactly
		$\mathbb{E}[(x_0-x_t)/t\mid x_t]$, allowing the stochastic reverse
		dynamics to be learned by ordinary quadratic regression on paired
		forward samples.
		
		\item
		A $\sqrt{t}$-scaled parameterization is introduced to remove the
		$\mathcal{O}(t^{-1/2})$ divergence of the direct regression target
		near $t=0$ while preserving the corresponding reverse dynamics.
		
		\item
		MVM is instantiated with Transformer-based generative models and
		evaluated on class-conditional ImageNet at $32\times32$ and
		$256\times256$. The stochastic reverse SDE and its deterministic
		probability-flow ODE are further compared under matched numerical
		budgets.
	\end{itemize}
	
	\section{RELATED WORK}
	
	\subsection{NOISE PREDICTION PARADIGMS}
	
	Noise prediction is a standard diffusion parameterization and is
	closely related to score matching \citep{11}, which estimates
	log-density gradients without evaluating normalized densities.
	Score-based generative models extend this idea to progressively
	perturbed distributions \citep{12,13}.
	
	DDPM \citep{2} uses a simple and effective noise parameterization.
	For a Gaussian perturbation process
	\begin{equation}
		x_t
		=
		\alpha_t x_0
		+
		\sigma_t\epsilon,
		\qquad
		\epsilon\sim\mathcal{N}(0,I),
	\end{equation}
	the network predicts $\epsilon$, which is analytically related to the
	score and therefore provides a convenient representation of reverse
	denoising dynamics.
	
	Continuous-time formulations generalize diffusion chains to SDEs
	\citep{5} and associate the reverse SDE with a probability-flow ODE
	sharing the same time-dependent marginals.
	
	Subsequent work improves training and sampling while retaining
	noise/score targets. Improved DDPM \citep{15} and Variational
	Diffusion Models \citep{16} refine objectives and variances, while
	Karras et al.~\citep{6} highlight the importance of preconditioning,
	noise-level sampling, loss weighting, and numerical solvers.
	
	For sampling, DDIM \citep{14} introduces deterministic implicit
	generation and DPM-Solver \citep{21} reduces network evaluations with
	high-order solvers. Classifier-free guidance \citep{20} controls the
	fidelity--diversity trade-off and is widely used in large-scale
	image-generation systems \citep{17,18,22}.
	
	Because equivalent parameterizations can induce different loss
	weightings and conditioning, clean-data and mixed signal--noise
	targets remain important alternatives.

	\subsection{DATA PREDICTION PARADIGMS}
	
	Another class of methods predicts the clean sample. Although $x_0$
	and $\epsilon$ are algebraically related under Gaussian perturbations,
	their regression errors are weighted differently across noise levels.
	
	Clean-data prediction is natural for generalized corruption.
	Cold Diffusion \citep{23} uses deterministic degradations such as
	blur, masking, and pixelation, while Soft Diffusion \citep{29}
	extends to linear corruption operators and predicts a clean image
	whose corrupted version matches the observation.
	
	Related ideas appear in discrete diffusion: Structured Denoising
	Diffusion Models \citep{27} predict clean discrete variables, and
	Vector Quantized Diffusion \citep{28} operates in a discrete latent
	space for text-to-image generation.
	
	Consistency Models \citep{30} likewise map states on the same
	probability-flow trajectory toward a common endpoint, enabling
	generation with few network evaluations. The relative effectiveness of data and noise prediction depends on the perturbation process and training weighting; neither target is universally superior, and each induces a different optimization
	geometry.

	\subsection{VELOCITY PREDICTION PARADIGMS}
	
	A third family combines signal and noise. Salimans and Ho
	\citep{25} introduced $v$-prediction for progressive distillation.
	Under a variance-preserving parameterization
	\begin{equation}
		x_t
		=
		\alpha_t x_0
		+
		\sigma_t\epsilon,
		\qquad
		\alpha_t^2+\sigma_t^2=1,
	\end{equation}
	the corresponding target can be written as $v_t	=\alpha_t\epsilon-\sigma_t x_0$.
	
	This target combines clean-data and noise information in one
	regression variable and improves numerical stability for progressive
	distillation.
	
	A conceptually different family models probability transport through
	neural vector fields. Continuous normalizing flows and neural ODEs
	\citep{38,39} use deterministic dynamics, while Flow Matching
	\citep{8} learns marginal vector fields along prescribed conditional
	probability paths.
	
	Rectified Flow \citep{10} progressively straightens transport paths
	to reduce discretization error and permit coarser integration.
	Optimal-transport extensions further improve path geometry
	\citep{40,41}, and rectified-flow Transformers scale well to
	high-resolution image synthesis \citep{42}.
	
	Deterministic flows and stochastic diffusions can describe the same
	density evolution. Score-based SDEs \citep{5} connect reverse-time
	SDEs with probability-flow ODEs, while Stochastic Interpolants
	\citep{9} relate transport equations, forward/backward
	Fokker--Planck equations, velocity fields, scores, and stochastic
	drifts in one framework.
	
	MVM differs from both diffusion $v$-prediction and the instantaneous
	vector fields of Flow Matching. For state $x_t$ and source $x_0$, the present study
	defines
	\begin{equation}
		v_{\mathrm{mean}}(x_t,t;x_0)
		=
		\frac{x_0-x_t}{t}.
	\end{equation}
	This is a finite-time average restoration velocity rather than an
	instantaneous path derivative.
	
	For the stochastic process considered in this paper, the conditional expectation
	\begin{equation}
		v^{*}(x_t,t)
		=
		\mathbb{E}
		\left[
		\left.
		\frac{x_0-x_t}{t}
		\right|
		x_t
		\right]
	\end{equation}
	is exactly the marginal reverse drift. Thus, the reverse SDE can be
	learned from one sample-wise quadratic target, distinct from diffusion
	$v$-prediction, deterministic Flow Matching, and general stochastic
	interpolants.

	\section{Mean Velocity Matching}
	
	This section derives MVM from probability-density evolution. Proofs of Theorems~1--6 are given in the appendix.
	
	\subsection{Single-Image Scenario}
	
	\subsubsection{Forward Process}
	
	For a single sample $x_0$, the initial density is a Dirac $\delta$ function centered at $x_0$. The forward process should transform this density into a standard normal distribution at $t=T$. This evolution is analyzed through its Fourier transform.
	
	\begin{equation}
		\mathcal{F}\left( p\left( x,t \right) \right)
		=\hat{p}\left( k,t \right)
		=\int_{-\infty}^{+\infty}p\left( x,t \right)e^{ikx}\,dx
	\end{equation}
	
	Here, $p$ is the density, $k$ the angular frequency, $\mathcal{F}$ the Fourier transform, and $\hat{p}(k,t)$ the characteristic function.
	
	\textbf{Theorem 1:} If the probability density function is a $\delta$ function, its characteristic function is
	\begin{equation}
		\hat{p}(k)=e^{ikx_0}.
	\end{equation}
	
	\textbf{Theorem 2:} If the probability density function is a normal distribution with mean $\mu$ and variance $\sigma^2$, its characteristic function is
	\begin{equation}
		\hat{p}(k)=e^{ik\mu -\frac{1}{2}\sigma^2k^2}.
	\end{equation}
	
	\textbf{Theorem 3:} If the probability density function is a standard normal distribution, its characteristic function is
	\begin{equation}
		\hat{p}(k)=e^{-\frac{k^2}{2}}.
	\end{equation}
	
	Assume that the evolution of the characteristic function over time follows the characteristic function transformation formula below.
	\begin{equation}
		\hat{p}(k,t) = \hat{p}(k,0)e^{-\lambda(k)t}
	\end{equation}
	By Theorems~1 and~3, the endpoint conditions are
	$\hat{p}(k,0)=e^{ikx_0}$ and $\hat{p}(k,T)=e^{-k^2/2}$, which determine
	\begin{equation}
		\lambda(k)
		=
		\frac{k^2}{2T}
		+
		\frac{ikx_0}{T}.
	\end{equation}
	Substituting $\lambda(k)$ and comparing the result with the characteristic
	function in Theorem~2 gives
	\begin{equation}
		\begin{aligned}
			\ln\hat{p}(k,t)
			&=
			ikx_0\left(1-\frac{t}{T}\right)
			-\frac{k^2t}{2T}
			=
			ik\mu_t-\frac{1}{2}\sigma_t^2k^2, \\
			\mu_t
			&=
			x_0\left(1-\frac{t}{T}\right),
			\qquad
			\sigma_t^2=\frac{t}{T}.
		\end{aligned}
	\end{equation}
	Consequently, the forward process is
	\begin{equation}
		x_t
		=
		\left(1-\frac{t}{T}\right)x_0
		+
		\sqrt{\frac{t}{T}}\,\varepsilon,
		\qquad
		\varepsilon\sim\mathcal{N}(0,1).
	\end{equation}
	
	\subsubsection{Reverse Process}
	The forward characteristic function is:
	\begin{equation}
		\ln \hat{p}(k,t) = ikx_0\left( 1 - \frac{t}{T} \right) - \frac{k^2t}{2T}
	\end{equation}
	Let reverse time be $\tau=T-t$.
	\begin{equation}
		\ln \hat{p}(k,\tau) = ikx_0\left( 1 - \frac{T - \tau}{T} \right) - \frac{k^2(T - \tau)}{2T} = ikx_0\left( \frac{\tau}{T} \right) - \frac{k^2}{2} + \frac{k^2\tau}{2T}
	\end{equation}
	Differentiating with respect to $\tau$ and using
	$\partial_{\tau}\hat{p}
	=\hat{p}\,\partial_{\tau}\ln\hat{p}$ gives
	\begin{equation}
		\begin{aligned}
			\frac{\partial\hat{p}}{\partial\tau}
			&=
			\hat{p}
			\left(
			ik\frac{x_0}{T}+\frac{k^2}{2T}
			\right) 
			&=
			\underbrace{ik\frac{x_0}{T}\hat{p}}_{A}
			+\underbrace{\frac{k^2}{T}\hat{p}}_{B}
			-\underbrace{\frac{k^2}{2T}\hat{p}}_{C}.
		\end{aligned}
		\label{eq:fourier_tau_derivative}
	\end{equation}
	
	Using $\mathcal{F}^{-1}(ik\hat p)=-\partial_xp$ and
	$\mathcal{F}^{-1}(-k^2\hat p)=\partial_x^2p$, the three terms become
	$A=-\partial_x[(x_0/T)p]$, $B=-(1/T)\partial_x^2p$, and
	$C=(1/2T)\partial_x^2p$.
	
	\begin{equation}
		\therefore \frac{\partial p}{\partial \tau} = -\frac{\partial}{\partial x}\left( \frac{x_0}{T}p \right) - \frac{1}{T}\frac{\partial^2 p}{\partial x^2} + \frac{1}{2T}\frac{\partial^2 p}{\partial x^2}
	\end{equation}
	
	Since $\partial_x p=p\nabla_x\ln p$, the $B$ term can be absorbed into the drift:
	
	\begin{equation}
		\frac{\partial p}{\partial \tau} = -\frac{\partial}{\partial x}\left[ \left( \frac{x_0}{T} + \frac{1}{T}\nabla_x\ln p \right) p \right] + \frac{1}{2}\frac{\partial^2}{\partial x^2}\left( \frac{1}{T}p \right)
	\end{equation}
	
	\textbf{Theorem: Fokker-Planck Equation}\citep{43}
	
	For an SDE $dx=f(x,t)dt+g(t)dw$, the Fokker--Planck equation is:
	
	\begin{equation}
		\frac{\partial p}{\partial t} = -\frac{\partial}{\partial x}\left[ f(x,t)p \right] + \frac{1}{2}\frac{\partial^2}{\partial x^2}\left[ g(t)^2p \right]
	\end{equation}
	
	Thus $f=\frac{x_0}{T}+\frac{1}{T}\nabla_x\ln p$ and $g=\sqrt{1/T}$. Substituting the score identity into the SDE gives
	\begin{equation}
		\begin{aligned}
			dx
			&=
			\left(
			\frac{x_0}{T}
			+\frac{1}{T}\nabla_x\ln p
			\right)d\tau
			+\sqrt{\frac{1}{T}}\,dw,
			\qquad
			\nabla_x\ln p
			=
			-\frac{\varepsilon}{\sqrt{t/T}}
			\\[-1mm]
			&=
			\left(
			\frac{x_0}{T}
			-\frac{\varepsilon}{\sqrt{tT}}
			\right)d\tau
			+\sqrt{\frac{1}{T}}\,dw .
		\end{aligned}
		\label{eq:forward_sde_noise_form}
	\end{equation}
	
	From the forward process, $\frac{x_0}{T}-\frac{\varepsilon}{\sqrt{tT}}=\frac{x_0-x_t}{t}$.
	
	Defining $v(x_t,t)=\frac{x_0-x_t}{t}$ gives the reverse SDE:
	
	\begin{equation}
		dx = v(x_t,t)d\tau + \sqrt{\frac{1}{T}}dw
	\end{equation}
	
	\subsection{Multi-Image Scenario}
	
	\subsubsection{Forward Process}
	
	Let the probability density for $N$ images be:
	
	\begin{equation}
		p(x_0) = \frac{1}{N}\sum_{j=1}^N \delta\left( x_0 - x_0^{(j)} \right)
	\end{equation}
	
	where $x_i^{(1)}, x_i^{(2)}, \dots, x_i^{(N)}$, $i \in [0, 1]$, denote the images at $t=i$.
	
	For each sample, use the same forward process:
	
	\begin{equation}
		x_t^{(k)} = \left( 1 - \frac{t}{T} \right)x_0^{(k)} + \sqrt{\frac{t}{T}}\varepsilon
	\end{equation}
	
	When $t=T$, $x_T^{(k)} \sim \mathcal{N}(0,1)$.
	
	The marginal density is:
	
	\begin{equation}
		\begin{aligned}
			p(x_t) 
			&= \int_{-\infty}^{+\infty} p(x_t \mid x_0)\left( \frac{1}{N}\sum_{j=1}^N \delta\left( x_0 - x_0^{(j)} \right) \right) dx_0 
			&= \frac{1}{N}\sum_{j=1}^N p\left( x_t \mid x_0^{(j)} \right)
		\end{aligned}
	\end{equation}
	
	Each conditional component satisfies $p(x_t\mid x_0^{(j)})\sim\mathcal{N}((1-t/T)x_0^{(j)},t/T)$ and becomes $\mathcal{N}(0,1)$ at $t=T$; therefore the mixture $p(x_T)$ is also standard normal.
	
	Thus the same forward process maps the full data distribution to a standard normal distribution.
	
	\subsubsection{Reverse Process}
	
	\textbf{Theorem 4:} If the reverse process for a single image is formulated as
	\begin{equation}dx = v(x_t,t)d\tau + \sqrt{\frac{1}{T}}dw\end{equation}
	then the reverse process for multiple images is given by
	\begin{equation}dx = v^*(x_t,t)d\tau + \sqrt{\frac{1}{T}}dw\end{equation}
	where the respective velocity fields are defined as
	\begin{equation}v(x_t,t) = \frac{x_0 - x_t}{t}, \quad v^*(x_t,t) = \mathbb{E}_{x_0 \mid x_t}\left[ \frac{x_0 - x_t}{t} \right]\end{equation}
	
	\textbf{Theorem 5:} Let the loss function $L(\theta)$ be defined as
	
	\begin{equation}
		L(\theta) = \mathbb{E}_{x_0 \sim data, x_t \mid x_0}\left[ \left\| v_\theta(x_t) - \frac{x_0 - x_t}{t} \right\|^2 \right]
	\end{equation}
	
	and the loss function $L^*(\theta)$ be defined as
	
	\begin{equation}
		L^*(\theta) = \mathbb{E}_{x_0 \sim data, x_t \mid x_0}\left[ \left\| v_\theta(x_t) - \mathbb{E}_{x_0 \mid x_t}\left[ \frac{x_0 - x_t}{t} \right] \right\|^2 \right]
	\end{equation}
	
	Then, optimizing $L(\theta)$ is mathematically equivalent to optimizing $L^*(\theta)$.
	
	By Theorem 5, it is sufficient to optimize $L(\theta)$, yielding the sample-wise reverse-process loss:
	
	\begin{equation}
		\left\| v_\theta(x_t, t) - v(x_t, t) \right\|^2
	\end{equation}
	
	Although Theorem~5 establishes the equivalence of the two regression
	objectives, directly using the mean-velocity target may lead to numerical
	instability near $t=0$. From the forward process,
	\begin{equation}
		\begin{aligned}
			v(x_t,t)
			&=
			\frac{x_0-x_t}{t} 
			&=
			\frac{x_0}{T}
			-
			\frac{\varepsilon}{\sqrt{tT}}.
		\end{aligned}
	\end{equation}
	Therefore, the magnitude of the direct regression target grows as
	$\mathcal{O}(t^{-1/2})$ and becomes unbounded when $t$ approaches zero.
	
	To obtain a bounded training target, this paper introduces the scaled
	mean velocity
	\begin{equation}
		\begin{aligned}
			u(x_t,t)
			&=
			\sqrt{t}\,v(x_t,t) 
			&=
			\frac{x_0-x_t}{\sqrt{t}} 
			&=
			\frac{\sqrt{t}}{T}x_0
			-
			\frac{\varepsilon}{\sqrt{T}}.
		\end{aligned}
	\end{equation}
	\begin{equation}
		\lim_{t\rightarrow 0}u(x_t,t)
		=
		-\frac{\varepsilon}{\sqrt{T}},
	\end{equation}
	which remains finite. The practical training objective is therefore
	defined as
	\begin{equation}
		\widetilde{L}(\theta)
		=
		\mathbb{E}_{x_0\sim p_{\mathrm{data}},\,x_t\mid x_0}
		\left[
		\left\|
		u_\theta(x_t,t)
		-
		\frac{x_0-x_t}{\sqrt{t}}
		\right\|^2
		\right].
	\end{equation}
	During sampling, the mean-velocity field is recovered by
	\begin{equation}
		v_\theta(x_t,t)
		=
		\frac{u_\theta(x_t,t)}{\sqrt{t}}.
	\end{equation}
	Accordingly, the reverse SDE used for generation is
	\begin{equation}
		dx
		=
		\frac{u_\theta(x_t,t)}{\sqrt{t}}\,d\tau
		+
		\sqrt{\frac{1}{T}}\,dw.
	\end{equation}
	
	\subsection{Deterministic ODE Formulation}
	
	In addition to the stochastic reverse process, MVM admits a deterministic
	probability-flow ODE with the same marginal distributions \citep{5}.
	
	\textbf{Theorem 6:} Let $p_t(x)$ denote the marginal density of the MVM
	forward process. The probability-flow ODE corresponding to the reverse SDE is
	\begin{equation}
		dx=
		\left[
		v^*(x_t,t)-\frac{1}{2T}\nabla_{x_t}\log p_t(x_t)
		\right]d\tau .
	\end{equation}
	For the MVM perturbation path, the score and mean velocity satisfy
	\begin{equation}
		\nabla_{x_t}\log p_t(x_t)=(T-t)v^*(x_t,t)-x_t .
	\end{equation}
	Consequently, the deterministic reverse process can be written as
	\begin{equation}
		dx=
		\left[
		\frac{T+t}{2T}v^*(x_t,t)+\frac{x_t}{2T}
		\right]d\tau .
	\end{equation}
	Since the network predicts $u_\theta(x_t,t)=\sqrt{t}\,v^*(x_t,t)$, the
	practical ODE is
	\begin{equation}
		dx=
		\left[
		\frac{T+t}{2T\sqrt{t}}u_\theta(x_t,t)+\frac{x_t}{2T}
		\right]d\tau .
	\end{equation}

	\section{Experiments}
	The experiments are conducted on class-conditional ImageNet at resolutions of
	$32\times32$ and $256\times256$ \citep{45}. ImageNet-$32$ is used to
	evaluate the basic generation performance of MVM, while ImageNet-$256$
	examines its ability to generate higher-resolution images. For ImageNet-$32$,
	the reverse process is parameterized by a pixel-space DiT \citep{44}, with
	REPA introduced to accelerate training \citep{46}. For ImageNet-$256$, an
	RAE with a frozen DINOv3 encoder reduces the spatial resolution before a
	DDT-style Transformer performs latent generation \citep{47,48}; REPA is also
	employed. SA-Solver is used for sampling \citep{49}. Classifier-free guidance
	is used in the ImageNet-$32$ SDE--ODE comparison, while REPA internal guidance
	is employed for the reported ImageNet-$256$ result. The first experiment
	evaluates generation quality using FID \citep{50}. The second compares the
	FID--NFE curves of the reverse SDE and its ODE counterpart on ImageNet-$32$
	\citep{5}.

	\subsection{Main Generation Results}
	
	This paper first evaluates the generation quality of Mean Velocity Matching
	(MVM) on ImageNet at two representative resolutions, namely $32\times32$ and
	$256\times256$. Following common practice in generative modeling,
	Fr\'echet Inception Distance (FID) is adopted as the primary metric, where a
	lower value indicates that the generated distribution is closer to the real
	data distribution. The number of function evaluations (NFE) is additionally
	reported to characterize the computational cost of sampling. All FID values
	of MVM are computed from $50{,}000$ generated samples.
	
	The compared methods cover representative diffusion-, score-, flow-, GAN-,
	latent-diffusion-, and Transformer-based generative models. The results are
	summarized in Tables~\ref{tab:imagenet32_fid} and
	\ref{tab:imagenet256_fid}.
	
\begin{table}[t]  
	\centering  
	\begin{minipage}[t]{0.47\textwidth}  
		\centering  
		\caption{Generation results on ImageNet $32\times32$.  
			Lower FID and NFE are better.}  
		\label{tab:imagenet32_fid}  
		\vspace{2mm}  
		
		\begin{tabular}{lcc}  
			\toprule  
			\textbf{Method} & \textbf{FID $\downarrow$} & \textbf{NFE $\downarrow$} \\  
			\midrule  
			DDPM                  & 6.99  & 262 \\  
			Score Matching        & 5.68  & 178 \\  
			ScoreFlow             & 14.14 & 195 \\  
			FM w/ Diffusion       & 6.37  & 193 \\  
			FM w/ OT              & 5.02  & 122 \\  
			\midrule
			
			\begin{tabular}[c]{@{}l@{}} 
				MVM w/ ODE \\ 
				{\scriptsize (No Guidance)} 
			\end{tabular} 
			& 2.27 & 65 \\[1mm]
			
			\begin{tabular}[c]{@{}l@{}} 
				MVM w/ ODE \\ 
				{\scriptsize (CFG=1.20, GI=[0.60,0.80])} 
			\end{tabular} 
			& 2.04 & 65 \\[1mm]
			
			\begin{tabular}[c]{@{}l@{}} 
				MVM w/ SDE \\ 
				{\scriptsize (No Guidance)} 
			\end{tabular} 
			& 2.26 & 65 \\[1mm]
			
			\begin{tabular}[c]{@{}l@{}} 
				\textbf{MVM w/ SDE} \\ 
				{\scriptsize (CFG=1.20, GI=[0.60,0.80])} 
			\end{tabular} 
			& \textbf{1.93} & \textbf{65} \\  
			
			\bottomrule  
		\end{tabular}  
		
	\end{minipage} 
	\hfill 
	\begin{minipage}[t]{0.47\textwidth}  
		\centering  
		\caption{Generation results on ImageNet $256\times256$.  
			Lower FID and NFE are better.}  
		\label{tab:imagenet256_fid}  
		\vspace{2mm}  
		
		\begin{tabular}{lcc}  
			\toprule  
			\textbf{Method} & \textbf{FID $\downarrow$} & \textbf{NFE $\downarrow$} \\  
			\midrule  
			ADM                   & 10.94 & 250 \\  
			StyleGAN-XL           & 2.30  & - \\  
			DiT-XL/2-G            & 2.27  & 250 \\  
			BigGAN-deep           & 6.95  & - \\  
			LDM-4-G               & 3.60  & 250 \\  
			\midrule
			
			\begin{tabular}[c]{@{}l@{}} 
				MVM w/ ODE \\ 
				{\scriptsize (No Guidance)} 
			\end{tabular} 
			& 2.40 & 90 \\[1mm]
			
			\begin{tabular}[c]{@{}l@{}} 
				MVM w/ ODE \\ 
				{\scriptsize (IG=1.45, GI=[0,1])} 
			\end{tabular} 
			& 2.16 & 90 \\[1mm]
			
			\begin{tabular}[c]{@{}l@{}} 
				MVM w/ SDE \\ 
				{\scriptsize (No Guidance)} 
			\end{tabular} 
			& 2.28 & 90 \\[1mm]
			
			\begin{tabular}[c]{@{}l@{}} 
				\textbf{MVM w/ SDE} \\ 
				{\scriptsize (IG=1.45, GI=[0,1])} 
			\end{tabular} 
			& \textbf{2.07} & \textbf{90} \\  
			
			\bottomrule  
		\end{tabular}  
		
		\vspace{1mm}  
	\end{minipage}  
\end{table}
	On ImageNet $32\times32$, MVM achieves an FID of $\MVMImageNetThirtyTwoFID$ with $\MVMImageNetThirtyTwoNFE$
	function evaluations using reverse-SDE sampling. As shown in
	Table~\ref{tab:imagenet32_fid}, the proposed method substantially improves
	upon the reported diffusion-, score-, and flow-based baselines while requiring
	fewer function evaluations.
	
	The present study further evaluates MVM on the more challenging
	class-conditional ImageNet $256\times256$ benchmark. As shown in
	Table~\ref{tab:imagenet256_fid}, MVM obtains an FID of
	$\MVMImageNetTwoFiftySixFID$ using $\MVMImageNetTwoFiftySixNFE$ function
	evaluations. This result demonstrates that the proposed mean-velocity
	formulation remains effective when scaling from low-resolution generation to
	higher-resolution class-conditional image synthesis.
	
	\subsection{Comparison of SDE and ODE Sampling}
	
	\begin{wrapfigure}[15]{r}{0.40\textwidth}
		\vspace{-0pt}
		\centering
		\includegraphics[width=\linewidth]
		{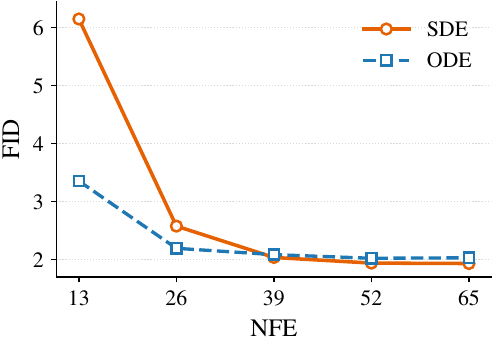}
		\caption{FID--NFE curves on ImageNet $32\times32$. Lower FID is better.}
		\label{fig:sde_ode_fid}
		\vspace{-0pt}
	\end{wrapfigure}
	
	The second experiment compares the stochastic reverse SDE with the
	deterministic probability-flow ODE on ImageNet $32\times32$. Both samplers use
	the same epoch-200 EMA checkpoint, initial Gaussian seeds, class labels,
	time-discretization rule, predictor--corrector orders, classifier-free-guidance
	configuration, number of generated images, and matched NFE. SA-Solver uses
	Adams predictor--corrector orders P3/C4, CFG scale $1.20$, and a guidance
	interval of $[0.60,0.80]$ measured in log-SNR progress. The ODE uses
	$\tau(t)=0$, whereas the SDE uses
	$\tau(t)=\rho\sqrt{(1-t)/(1+t)}$ with $\rho=0.75$. 
	
	At NFE values of $13$ and $26$, the probability-flow ODE obtains FID values
	of $3.356$ and $2.197$, respectively, outperforming the reverse SDE values of
	$6.148$ and $2.579$. The ordering reverses as the computational budget
	increases. At NFE values of $39$, $52$, and $65$, the reverse SDE achieves
	FID values of $2.039$, $1.942$, and $1.935$, respectively, whereas the ODE
	obtains $2.088$, $2.025$, and $2.034$. As shown in
	Figure~\ref{fig:sde_ode_fid}, the ODE is preferable under a strict low-NFE
	budget, while the stochastic reverse process yields lower FID from NFE $=39$
	onward. The best observed SDE result is $1.9349$ at NFE $=65$, whereas the
	best ODE result is $2.0253$ at NFE $=52$. Therefore, sufficient numerical
	resolution is particularly important for the stochastic reverse dynamics,
	which ultimately attain better generation quality than the corresponding
	deterministic flow.
	
	\section{Conclusion}
	
	This paper introduces Mean Velocity Matching (MVM), a generative framework
	that constructs a tractable diffusion path through Fourier transform
	and learns the reverse dynamics using a single mean-velocity field. The
	resulting formulation supports both stochastic reverse-SDE and deterministic
	probability-flow ODE sampling while preserving the same forward process.
	Experiments on ImageNet $32\times32$ and $256\times256$ demonstrate strong
	generation quality, achieving FID values of $\MVMImageNetThirtyTwoFID$ and
	$\MVMImageNetTwoFiftySixFID$, respectively.
	The SDE--ODE comparison further shows that the ODE is advantageous at very
	low NFE, whereas the SDE attains lower FID when more function evaluations are
	available. These results establish MVM as an effective and scalable framework
	for image generation.
	
	
	\section*{AI Use Statement}
	
	Generative AI tools were used to assist with literature discovery and to improve the clarity and readability of the manuscript. 
	All references identified with AI assistance were independently verified by the authors, and all AI-assisted edits were reviewed by the authors, who take full responsibility for the final content.
	
	\bibliographystyle{iclr2027_conference}
	\bibliography{references}
	
	\appendix
	
	\section{Proofs of Theorems}
	\label{app:theorem_proofs}
	
	This appendix contains the complete proofs of Theorems~1--6 stated in
	the main text. Throughout the appendix, the Fourier transform is defined by
	\begin{equation}
		\widehat{p}(k)
		=
		\mathcal{F}[p](k)
		=
		\int_{-\infty}^{+\infty}p(x)e^{ikx}\,dx.
		\label{eq:app_fourier_definition}
	\end{equation}
	All differentiations under the integral sign and integrations by parts are
	understood under the standard integrability and smoothness conditions. These
	conditions hold for the Gaussian transition densities used by MVM for every
	$t>0$.
	
	\subsection{Proof of Theorem 1}
	
	For a deterministic sample $x_0$, the associated probability measure is the
	Dirac measure $\delta(x-x_0)$. Applying the defining sifting property of the
	Dirac delta to Equation~\eqref{eq:app_fourier_definition} gives
	\begin{equation}
		\begin{aligned}
			\widehat{p}(k)
			&=
			\int_{-\infty}^{+\infty}
			\delta(x-x_0)e^{ikx}\,dx \\
			&=
			e^{ikx_0}.
		\end{aligned}
		\label{eq:app_dirac_characteristic}
	\end{equation}
	Hence, the characteristic function of a point mass at $x_0$ is
	$e^{ikx_0}$. This proves Theorem~1. \hfill $\square$
	
	\subsection{Proof of Theorem 2}
	
	Let $X\sim\mathcal{N}(\mu,\sigma^2)$ and write
	\begin{equation}
		X=\mu+\sigma Z,
		\qquad
		Z\sim\mathcal{N}(0,1).
		\label{eq:app_gaussian_standardization}
	\end{equation}
	It is sufficient first to obtain the characteristic function of $Z$. Denote
	its density by $\phi(z)=(2\pi)^{-1/2}e^{-z^2/2}$ and define
	\begin{equation}
		I(a)
		=
		\mathbb{E}[e^{iaZ}]
		=
		\int_{-\infty}^{+\infty}e^{iaz}\phi(z)\,dz.
		\label{eq:app_standard_normal_cf_integral}
	\end{equation}
	Because $|z|\phi(z)$ is integrable, differentiation under the integral sign is
	valid. Since $\phi'(z)=-z\phi(z)$, integration by parts yields
	\begin{equation}
		\begin{aligned}
			I'(a)
			&=
			i\int_{-\infty}^{+\infty}z e^{iaz}\phi(z)\,dz \\
			&=
			-i\int_{-\infty}^{+\infty}e^{iaz}\phi'(z)\,dz \\
			&=
			-i\left(
			\left[e^{iaz}\phi(z)\right]_{-\infty}^{+\infty}
			-ia\int_{-\infty}^{+\infty}e^{iaz}\phi(z)\,dz
			\right) \\
			&=
			-aI(a).
		\end{aligned}
		\label{eq:app_gaussian_cf_ode}
	\end{equation}
	The boundary term in Equation~\eqref{eq:app_gaussian_cf_ode} vanishes because
	$\phi(z)$ decays exponentially. Moreover,
	\begin{equation}
		I(0)
		=
		\int_{-\infty}^{+\infty}\phi(z)\,dz
		=1.
		\label{eq:app_gaussian_cf_initial_condition}
	\end{equation}
	Solving the ordinary differential equation in
	Equation~\eqref{eq:app_gaussian_cf_ode} subject to
	Equation~\eqref{eq:app_gaussian_cf_initial_condition} gives
	\begin{equation}
		I(a)=e^{-a^2/2}.
		\label{eq:app_standard_normal_cf}
	\end{equation}
	Consequently, the characteristic function of $X$ is
	\begin{equation}
		\begin{aligned}
			\widehat{p}(k)
			&=
			\mathbb{E}[e^{ikX}] \\
			&=
			\mathbb{E}[e^{ik(\mu+\sigma Z)}] \\
			&=
			e^{ik\mu}I(k\sigma) \\
			&=
			\exp\!\left(ik\mu-\frac{1}{2}\sigma^2k^2\right).
		\end{aligned}
		\label{eq:app_general_gaussian_cf}
	\end{equation}
	This proves Theorem~2. \hfill $\square$
	
	\subsection{Proof of Theorem 3}
	
	A standard normal distribution is the special case $\mu=0$ and
	$\sigma^2=1$ of Theorem~2. Substitution into
	Equation~\eqref{eq:app_general_gaussian_cf} gives
	\begin{equation}
		\widehat{p}(k)
		=
		\exp\!\left(ik\cdot0-\frac{1}{2}k^2\right)
		=
		e^{-k^2/2}.
		\label{eq:app_standard_gaussian_cf}
	\end{equation}
	This proves Theorem~3. \hfill $\square$
	
	\subsection{Proof of Theorem 4}
	
	Let $\pi(dx_0)$ denote the data distribution. For a fixed clean sample $x_0$,
	let $q_t(x\mid x_0)$ be the forward conditional density, and introduce reverse
	time and the corresponding conditional density by
	\begin{equation}
		\tau=T-t,
		\qquad
		r_\tau(x\mid x_0)=q_{T-\tau}(x\mid x_0).
		\label{eq:app_reverse_time_density}
	\end{equation}
	The single-image reverse process assumed in Theorem~4 is
	\begin{equation}
		dX_\tau
		=
		v(X_\tau,t;X_0)\,d\tau
		+\frac{1}{\sqrt{T}}\,d\overline{W}_\tau,
		\qquad
		v(x,t;x_0)=\frac{x_0-x}{t},
		\qquad
		t=T-\tau.
		\label{eq:app_single_reverse_sde}
	\end{equation}
	Therefore, for every fixed $x_0$, its conditional reverse density satisfies
	the Fokker--Planck equation
	\begin{equation}
		\partial_\tau r_\tau(x\mid x_0)
		=
		-\partial_x\!\left[
		v(x,t;x_0)r_\tau(x\mid x_0)
		\right]
		+\frac{1}{2T}\partial_x^2r_\tau(x\mid x_0).
		\label{eq:app_conditional_reverse_fpe}
	\end{equation}
	The marginal density in reverse time is obtained by averaging the conditional
	density over the data distribution:
	\begin{equation}
		r_\tau(x)
		=
		\int r_\tau(x\mid x_0)\,\pi(dx_0).
		\label{eq:app_reverse_marginal_density}
	\end{equation}
	Differentiating Equation~\eqref{eq:app_reverse_marginal_density} with respect
	to $\tau$, substituting Equation~\eqref{eq:app_conditional_reverse_fpe}, and
	interchanging differentiation and integration gives
	\begin{equation}
		\begin{aligned}
			\partial_\tau r_\tau(x)
			&=
			\int \partial_\tau r_\tau(x\mid x_0)\,\pi(dx_0) \\
			&=
			-\partial_x\!\left[
			\int v(x,t;x_0)r_\tau(x\mid x_0)\,\pi(dx_0)
			\right]
			+\frac{1}{2T}\partial_x^2
			\int r_\tau(x\mid x_0)\,\pi(dx_0) \\
			&=
			-\partial_x\!\left[
			\int v(x,t;x_0)r_\tau(x\mid x_0)\,\pi(dx_0)
			\right]
			+\frac{1}{2T}\partial_x^2r_\tau(x).
		\end{aligned}
		\label{eq:app_marginalized_reverse_fpe}
	\end{equation}
	Whenever $r_\tau(x)>0$, Bayes' rule gives the posterior distribution of the
	clean sample conditioned on the current state:
	\begin{equation}
		\pi(dx_0\mid x,\tau)
		=
		\frac{r_\tau(x\mid x_0)\,\pi(dx_0)}{r_\tau(x)}.
		\label{eq:app_bayes_posterior}
	\end{equation}
	Using Equation~\eqref{eq:app_bayes_posterior}, the drift integral in
	Equation~\eqref{eq:app_marginalized_reverse_fpe} can be rewritten as
	\begin{equation}
		\begin{aligned}
			\int v(x,t;x_0)r_\tau(x\mid x_0)\,\pi(dx_0)
			&=
			r_\tau(x)\int v(x,t;x_0)\,\pi(dx_0\mid x,\tau) \\
			&=
			r_\tau(x)\,
			\mathbb{E}\!\left[
			\left.v(x,t;X_0)\right|X_t=x
			\right].
		\end{aligned}
		\label{eq:app_posterior_drift_average}
	\end{equation}
	Define the marginal reverse drift as
	\begin{equation}
		\begin{aligned}
			v^*(x,t)
			&:=
			\mathbb{E}\!\left[
			\left.v(x,t;X_0)\right|X_t=x
			\right] \\
			&=
			\mathbb{E}\!\left[
			\left.\frac{X_0-x}{t}\right|X_t=x
			\right].
		\end{aligned}
		\label{eq:app_marginal_reverse_drift}
	\end{equation}
	Substituting Equations~\eqref{eq:app_posterior_drift_average} and
	\eqref{eq:app_marginal_reverse_drift} into
	Equation~\eqref{eq:app_marginalized_reverse_fpe} yields
	\begin{equation}
		\partial_\tau r_\tau(x)
		=
		-\partial_x\!\left[v^*(x,t)r_\tau(x)\right]
		+\frac{1}{2T}\partial_x^2r_\tau(x).
		\label{eq:app_marginal_reverse_fpe}
	\end{equation}
	Equation~\eqref{eq:app_marginal_reverse_fpe} is precisely the
	Fokker--Planck equation associated with
	\begin{equation}
		dX_\tau
		=
		v^*(X_\tau,t)\,d\tau
		+\frac{1}{\sqrt{T}}\,d\overline{W}_\tau,
		\qquad
		t=T-\tau.
		\label{eq:app_marginal_reverse_sde}
	\end{equation}
	Thus, marginalizing the single-image reverse dynamics replaces the
	sample-dependent velocity by its posterior conditional expectation while
	leaving the diffusion coefficient unchanged. This proves Theorem~4.
	\hfill $\square$
	
	\subsection{Proof of Theorem 5}
	
	For a fixed $t>0$, define the observable sample-wise target and its conditional
	mean by
	\begin{equation}
		Y_t
		:=
		\frac{X_0-X_t}{t},
		\qquad
		m_t(X_t)
		:=
		\mathbb{E}[Y_t\mid X_t]
		=
		v^*(X_t,t).
		\label{eq:app_target_and_conditional_mean}
	\end{equation}
	With this notation, the two population losses in Theorem~5 are
	\begin{equation}
		L(\theta)
		=
		\mathbb{E}\!\left[
		\left\|v_\theta(X_t,t)-Y_t\right\|^2
		\right]
		\label{eq:app_samplewise_loss}
	\end{equation}
	and
	\begin{equation}
		L^*(\theta)
		=
		\mathbb{E}\!\left[
		\left\|v_\theta(X_t,t)-m_t(X_t)\right\|^2
		\right].
		\label{eq:app_conditional_mean_loss}
	\end{equation}
	Add and subtract $m_t(X_t)$ inside
	Equation~\eqref{eq:app_samplewise_loss}. Expanding the squared Euclidean norm
	gives
	\begin{equation}
		\begin{aligned}
			L(\theta)
			&=
			\mathbb{E}\!\left[
			\left\|
			v_\theta(X_t,t)-m_t(X_t)+m_t(X_t)-Y_t
			\right\|^2
			\right] \\
			&=
			\mathbb{E}\!\left[
			\left\|v_\theta(X_t,t)-m_t(X_t)\right\|^2
			\right]
			+
			\mathbb{E}\!\left[
			\left\|Y_t-m_t(X_t)\right\|^2
			\right] \\
			&\quad
			+2\mathbb{E}\!\left[
			\bigl(v_\theta(X_t,t)-m_t(X_t)\bigr)^\top
			\bigl(m_t(X_t)-Y_t\bigr)
			\right].
		\end{aligned}
		\label{eq:app_loss_expansion}
	\end{equation}
	The final cross-term in Equation~\eqref{eq:app_loss_expansion} is zero. Indeed,
	the tower property of conditional expectation and the definition of
	$m_t(X_t)$ imply
	\begin{equation}
		\begin{aligned}
			&\mathbb{E}\!\left[
			\bigl(v_\theta(X_t,t)-m_t(X_t)\bigr)^\top
			\bigl(m_t(X_t)-Y_t\bigr)
			\right] \\
			&\quad=
			\mathbb{E}\!\left[
			\bigl(v_\theta(X_t,t)-m_t(X_t)\bigr)^\top
			\mathbb{E}\!\left[
			\left.m_t(X_t)-Y_t\right|X_t
			\right]
			\right] \\
			&\quad=
			\mathbb{E}\!\left[
			\bigl(v_\theta(X_t,t)-m_t(X_t)\bigr)^\top
			\bigl(m_t(X_t)-\mathbb{E}[Y_t\mid X_t]\bigr)
			\right] \\
			&\quad=0.
		\end{aligned}
		\label{eq:app_cross_term_zero}
	\end{equation}
	Combining Equations~\eqref{eq:app_conditional_mean_loss},
	\eqref{eq:app_loss_expansion}, and \eqref{eq:app_cross_term_zero} gives the
	conditional-expectation Pythagorean identity
	\begin{equation}
		L(\theta)
		=
		L^*(\theta)
		+
		\underbrace{
			\mathbb{E}\!\left[
			\left\|Y_t-m_t(X_t)\right\|^2
			\right]
		}_{C_t}.
		\label{eq:app_pythagorean_identity}
	\end{equation}
	The term $C_t$ is independent of $\theta$. Therefore,
	\begin{equation}
		\operatorname*{arg\,min}_{\theta}L(\theta)
		=
		\operatorname*{arg\,min}_{\theta}L^*(\theta),
		\qquad
		\nabla_\theta L(\theta)
		=
		\nabla_\theta L^*(\theta)
		\label{eq:app_loss_equivalence}
	\end{equation}
	whenever the gradients exist. Hence, regression to the sample-wise target
	$(X_0-X_t)/t$ and regression to its conditional mean have identical population
	minimizers.
	
	If training also samples $t$ from a distribution $\rho$, the same argument is
	applied after conditioning on the joint input $(X_t,t)$. In that case,
	\begin{equation}
		\begin{aligned}
			&\mathbb{E}\!\left[
			\left\|v_\theta(X_t,t)-Y_t\right\|^2
			\right] \\
			&\quad=
			\mathbb{E}\!\left[
			\left\|v_\theta(X_t,t)-\mathbb{E}[Y_t\mid X_t,t]\right\|^2
			\right]
			+
			\mathbb{E}\!\left[
			\left\|Y_t-\mathbb{E}[Y_t\mid X_t,t]\right\|^2
			\right].
		\end{aligned}
		\label{eq:app_random_time_loss_decomposition}
	\end{equation}
	The second term in Equation~\eqref{eq:app_random_time_loss_decomposition} is
	again independent of $\theta$, so the equivalence also holds for the
	time-averaged training objective. This proves Theorem~5. \hfill $\square$
	
	\subsection{Proof of Theorem 6}
	\label{app:proof_theorem6}
	
	For the MVM forward process, the conditional density is
	\begin{equation}
		p_t(x\mid x_0)
		=
		\mathcal{N}\!\left(
		x;\left(1-\frac{t}{T}\right)x_0,\frac{t}{T}I
		\right),
		\qquad 0<t\leq T.
		\label{eq:app_t6_forward_kernel}
	\end{equation}
	Here, $x$ denotes a spatial variable, and the reverse time is
	$\tau=T-t$. By Theorem~4, the marginal reverse SDE is
	\begin{equation}
		dx
		=
		v^*(x_t,t)\,d\tau
		+\frac{1}{\sqrt{T}}\,d\overline{W}_\tau.
		\label{eq:app_t6_reverse_sde}
	\end{equation}
	Its marginal density satisfies the Fokker--Planck equation
	\begin{equation}
		\partial_\tau p_t(x)
		=
		-\nabla_x\!\cdot\!\left[v^*(x,t)p_t(x)\right]
		+\frac{1}{2T}\Delta_x p_t(x),
		\label{eq:app_t6_reverse_fpe}
	\end{equation}
	where $\partial_\tau p_t(x)$ means differentiation of
	$p_{T-\tau}(x)$ with $x$ fixed.
	
	Using $\nabla_x p_t=p_t\nabla_x\log p_t$, this equation becomes
	\begin{equation}
		\begin{aligned}
			\partial_\tau p_t(x)
			&=
			-\nabla_x\!\cdot\!\left[v^*(x,t)p_t(x)\right]
			+\frac{1}{2T}\nabla_x\!\cdot
			\left[p_t(x)\nabla_x\log p_t(x)\right] \\
			&=
			-\nabla_x\!\cdot
			\left[
			\left(
			v^*(x,t)-\frac{1}{2T}\nabla_x\log p_t(x)
			\right)p_t(x)
			\right].
		\end{aligned}
		\label{eq:app_t6_fpe_continuity}
	\end{equation}
	This is the continuity equation associated with the
	probability-flow ODE
	\begin{equation}
		dx
		=
		\left[
		v^*(x_t,t)-\frac{1}{2T}
		\nabla_{x_t}\log p_t(x_t)
		\right]d\tau.
		\label{eq:app_t6_probability_flow}
	\end{equation}
	Thus, the prescribed marginal density path $p_t$ satisfies
	the density equation of this deterministic flow, with the
	same Gaussian initialization $p_T=\mathcal{N}(0,I)$.
	
	It remains to express the score using the mean velocity.
	The conditional Gaussian score is
	\begin{equation}
		\nabla_x\log p_t(x\mid x_0)
		=
		-\frac{T}{t}
		\left[
		x-\left(1-\frac{t}{T}\right)x_0
		\right].
		\label{eq:app_t6_conditional_score}
	\end{equation}
	Since
	$p_t(x)=\mathbb{E}_{x_0\sim\mathrm{data}}
	[p_t(x\mid x_0)]$,
	differentiation and Bayes' rule give
	\begin{equation}
		\begin{aligned}
			\nabla_x\log p_t(x)
			&=
			\frac{
				\mathbb{E}_{x_0\sim\mathrm{data}}
				\left[
				p_t(x\mid x_0)\nabla_x\log p_t(x\mid x_0)
				\right]
			}{p_t(x)} \\
			&=
			\mathbb{E}\!\left[
			\left.\nabla_x\log p_t(x\mid x_0)
			\right|x_t=x
			\right] \\
			&=
			-\frac{T}{t}
			\left[
			x-\left(1-\frac{t}{T}\right)
			\mathbb{E}[x_0\mid x_t=x]
			\right].
		\end{aligned}
		\label{eq:app_t6_score_identity}
	\end{equation}
	By the definition of the exact marginal mean velocity,
	\begin{equation}
		\begin{aligned}
			v^*(x,t)
			&=
			\mathbb{E}\!\left[
			\left.\frac{x_0-x}{t}\right|x_t=x
			\right]
			=
			\frac{\mathbb{E}[x_0\mid x_t=x]-x}{t}, \\
			\mathbb{E}[x_0\mid x_t=x]
			&=x+t\,v^*(x,t).
		\end{aligned}
		\label{eq:app_t6_velocity_mean_relation}
	\end{equation}
	Substituting this relation into
	Equation~\eqref{eq:app_t6_score_identity} yields
	\begin{equation}
		\begin{aligned}
			\nabla_x\log p_t(x)
			&=
			-\frac{T}{t}
			\left[
			x-\left(1-\frac{t}{T}\right)
			\left(x+t\,v^*(x,t)\right)
			\right] \\
			&=
			-\frac{T}{t}
			\left[
			\frac{t}{T}x
			-\frac{t(T-t)}{T}v^*(x,t)
			\right] \\
			&=(T-t)v^*(x,t)-x.
		\end{aligned}
		\label{eq:app_t6_score_velocity_relation}
	\end{equation}
	
	Finally, evaluating this identity at $x=x_t$ and substituting
	it into Equation~\eqref{eq:app_t6_probability_flow} gives
	\begin{equation}
		\begin{aligned}
			dx
			&=
			\left[
			v^*(x_t,t)
			-\frac{(T-t)v^*(x_t,t)-x_t}{2T}
			\right]d\tau \\
			&=
			\left[
			\frac{T+t}{2T}v^*(x_t,t)
			+\frac{x_t}{2T}
			\right]d\tau.
		\end{aligned}
		\label{eq:app_t6_final_ode}
	\end{equation}
	Replacing $v^*(x_t,t)$ by the learned approximation
	$u_\theta(x_t,t)/\sqrt{t}$ gives the practical ODE
	\begin{equation}
		dx
		=
		\left[
		\frac{T+t}{2T\sqrt{t}}u_\theta(x_t,t)
		+\frac{x_t}{2T}
		\right]d\tau.
		\label{eq:app_t6_practical_ode}
	\end{equation}
	This completes the derivation. \hfill $\square$

	\begin{figure}[p]
		\centering
		\includegraphics[
		height=0.78\textheight,
		keepaspectratio
		]{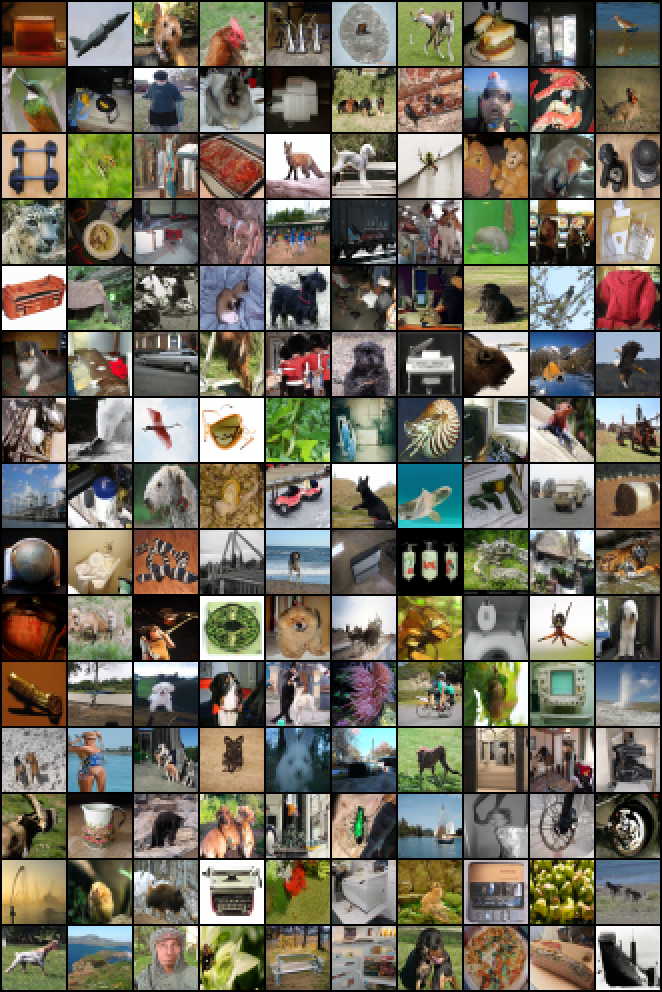}
		\caption{Uncurated class-conditional samples generated by MVM on
			ImageNet $32\times32$. The epoch-200 EMA checkpoint is sampled
			using the stochastic SA-Solver with 65 NFE.}
		\label{fig:appendix_imagenet32_samples}
	\end{figure}
	
	\clearpage
	
	\begin{figure}[p]
		\centering
		\includegraphics[
		height=0.78\textheight,
		keepaspectratio
		]{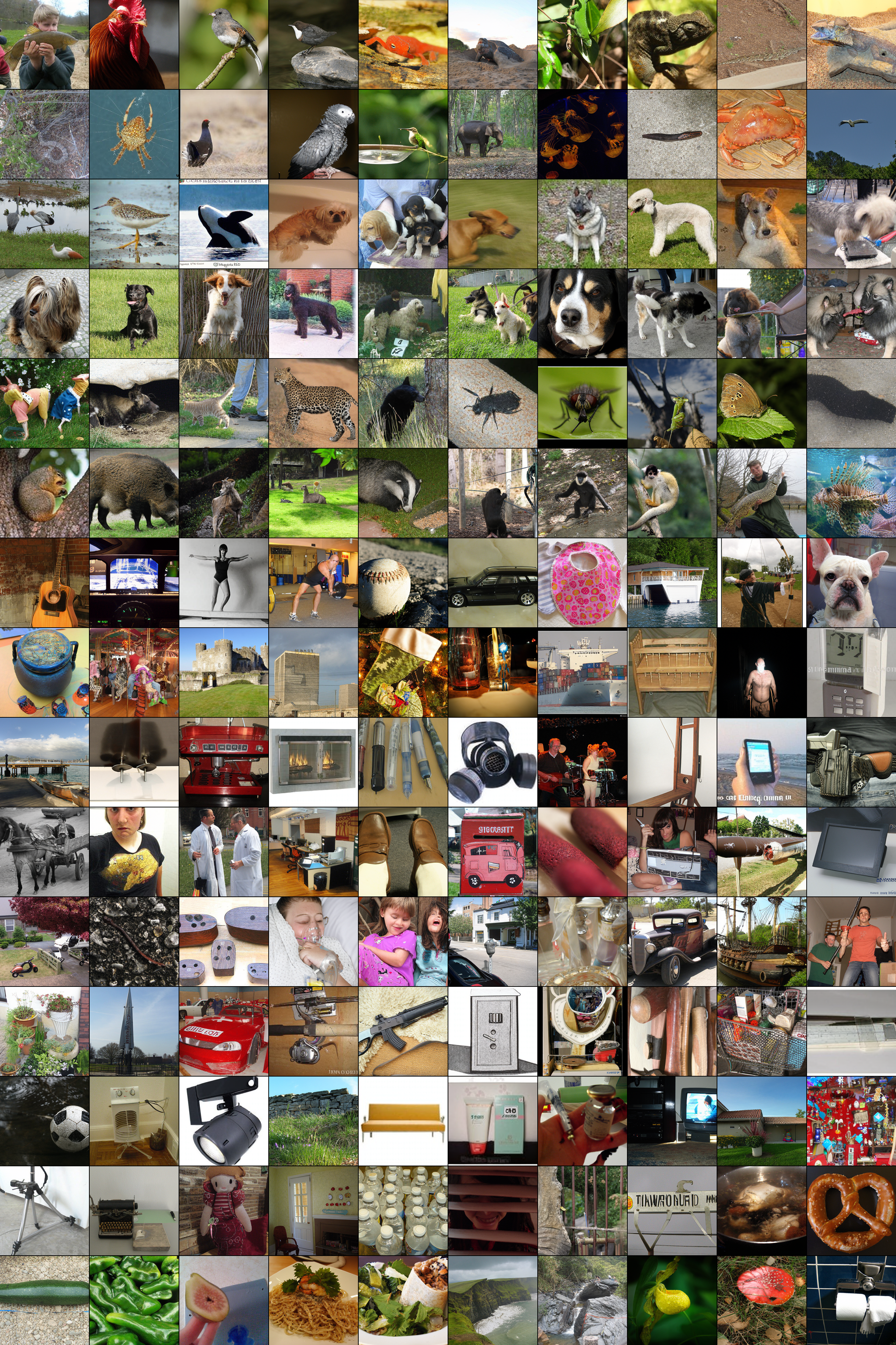}
		\caption{Uncurated class-conditional samples generated by MVM on ImageNet $256\times256$. The epoch-80 EMA checkpoint is sampled using the stochastic SA-Solver with 90 NFE.}
		\label{fig:appendix_imagenet256_samples}
	\end{figure}
	
	\clearpage
\end{document}